\documentclass{llncs}

\usepackage{psfig}

\begin{document}

\title{On the accuracy and running time of GSAT}
\author{Deborah East and Miroslaw Truszczynski\\
        Department of Computer Science}
\institute{ University of Kentucky\\
        Lexington KY 40506-0046\\
        \email{deast|mirek@cs.uky.edu}}

\maketitle
\begin{abstract}
Randomized algorithms for deciding satisfiability were shown to be
effective in solving problems with thousands of variables. However, 
these algorithms are not complete. That is, they provide no
guarantee that a satisfying assignment, if one exists, will be found.
Thus, when studying randomized algorithms, there are two important 
characteristics that need to be considered: the running time and,
even more importantly, the accuracy --- a measure of likelihood that a
satisfying assignment will be found, provided one exists. In fact, 
we argue that without a reference to the accuracy, the notion of the
running time for randomized algorithms is not well-defined. In this
paper, we introduce a formal notion of accuracy. We use it to define 
a concept of the running time. We use both notions to study the
random walk strategy GSAT algorithm. We investigate the dependence of
accuracy on properties of input formulas such as clause-to-variable 
ratio and the number of satisfying assignments. We demonstrate that 
the running time of GSAT grows exponentially in the number of 
variables of the input formula for randomly generated 3-CNF formulas 
and for the formulas encoding 3- and 4-colorability of graphs. 
\end{abstract}

\section{Introduction}
\label{intro}

The problem of deciding satisfiability of a boolean formula is 
extensively studied in computer science. It appears prominently,
as a prototypical NP-complete problem, in the investigations of 
computational complexity classes. It is studied by the automated 
theorem proving community. It is also of substantial interest 
to the AI community due to its applications in several areas including 
knowledge representation, diagnosis and planning.  

Deciding satisfiability of a boolean formula is an NP-complete
problem. Thus, it is unlikely that sound and complete algorithms 
running in polynomial time exist. However, recent years
brought several significant advances. First, fast (although, clearly,
still exponential in the worst case) implementations of the celebrated 
Davis-Putnam procedure \cite{dp60} were found. These implementations are
able to determine in a matter of seconds the satisfiability of critically
constrained CNF formulas with 300 variables and thousands of clauses 
\cite{dub96}. Second, several fast randomized algorithms were 
proposed and thoroughly studied \cite{slm92,skc96,sk93,msg97,spe96}. 
These algorithms randomly generate valuations and then apply some 
local improvement method in 
an attempt to reach a satisfying assignment. They are often very
fast but they provide no guarantee that, given a {\em satisfiable} 
formula, a satisfying assignment will be found. That is, randomized 
algorithms, while often fast, are not complete. Still, they were shown 
to be quite effective and solved several practical large-scale 
satisfiability problems \cite{sk92}.

One of the most extensively studied randomized algorithms recently is
GSAT \cite{slm92}. GSAT was shown to outperform the Davis-Putnam
procedure on randomly generated 3-CNF formulas from the crossover 
region \cite{slm92}. However, GSAT's performance on structured formulas
(encoding coloring and planning problems) was poorer \cite{skc96,sk93,skc94}. 
The basic GSAT algorithm would often become trapped within local minima 
and never reach a solution. To remedy this, several strategies for 
escaping from local minima were added to GSAT yielding its variants:
GSAT with averaging, GSAT with clause weighting, GSAT with random walk
strategy (RWS-GSAT), among others \cite{sk93,skc94}.
GSAT with random walk strategy was shown to perform especially well. 
These studies, while conducted on a wide range of classes of formulas 
rarely address a
critical issue of the {\em likelihood} that GSAT will find a satisfying
assignment, if one exists, and the running time is studied without
a reference to this likelihood. Notable exceptions are \cite{spe96}, where
RWS-GSAT is compared with a simulated annealing algorithm SASAT, 
and \cite{msg97}, where RSW-GSAT is compared to a tabu search method.

In this paper, we propose a systematic approach for studying the
quality of randomized algorithms. To this end, we introduce the concepts
of the {\em accuracy} and of the running time relative to the accuracy.
The accuracy measures how likely it is that a randomized algorithm
finds a satisfying assignment, assuming that the input formula is 
satisfiable. It is clear that the accuracy of GSAT (and any other similar 
randomized algorithm) grows as a function of time --- the longer we let 
the algorithm run, the better the chance that it will find a satisfying 
valuation (if one exists). In this paper, we present experimental results 
that allow us to quantify this intuition and get insights into the rate
of growth of the accuracy.

The notion of the running time of a randomized algorithm has not been 
rigorously studied. First, in most cases, a randomized algorithm has 
its running time determined by 
the choice of parameters that specify the number of random guesses,
the number of random steps in a local improvement process, etc. Second, 
in practical applications, randomized algorithms are often used in 
an interactive way. The algorithm is allowed to run until it finds a solution 
or the user decides not to wait any more, stops the execution, modifies 
the parameters of the algorithm or modifies the problem, and tries again. 
Finally, since randomized algorithms are not complete, they may make 
errors by not finding satisfying assignments when such assignments exist. 
Algorithms that are faster may be less accurate and the trade-off must
be taken into consideration \cite{spe96}.

It all points to the problems that arise when attempting to systematically 
study the running times of randomized algorithms and extrapolate their 
asymptotic behavior. In this paper, we define the concept of a {\em running
time relative to the accuracy}. The relative running time is,
intuitively, the time needed by a randomized algorithm to guarantee 
a postulated accuracy. We show in the paper that the relative running 
time is a useful performance measure for randomized satisfiability 
testing algorithms. In particular, we show that the running time of
GSAT relative to a prescribed accuracy grows {\em exponentially} with the
size of the problem.

Related work where the emphasis has been on fine tuning parameter
settings \cite{pw96,gw95} has shown somewhat different results in
regard to the increase in time as the size of the problems grow.  
The growth shown by \cite{pw96} is the retropective variation of
maxflips rather than the total number of flips.  The number of 
variables for the 3-CNF randomized instances reported \cite{gw95}
are  $50, 70, 100$.  Although our results are also limited by
the ability of complete algorithms to determine satisfiable instances,
we have results for $50, 100, \ldots, 400$ variable instances in the
crossover region.  The focus in our work is on maintaining accuracy 
as the size of the problems increase.

Second, we study the dependence of the accuracy and the relative running
time on the number of satisfying assignments that the input formula admits.  
Intuitively, the more satisfying assignments the input formula has, 
the better the chance that a randomized algorithm finds one of them, and 
the shorter the time needed to do so. Again, our results quantify these 
intuitions. We show that the performance of GSAT increases {\em exponentially}
with the growth in the number of satisfying assignments.

These results have interesting implications for the problem of
constructing sets of test cases for experimenting with satisfiability
algorithms. It is now commonly accepted that random $k$-CNF formulas from 
the cross-over region are ``difficult'' from the point of view of 
deciding their satisfiability. Consequently, they are good candidates
for testing satisfiability algorithms. These claims are based on the
studies of the performance of the Davis-Putnam procedure. Indeed, 
on average, it takes the most time to decide satisfiability of CNF formulas
randomly generated from the cross-over region. However, the suitability of 
formulas generated randomly from the cross-over region for the studies of the
performance of randomized algorithms is less clear. Our results
indicate that the performance of randomized algorithms critically depends
on the number of satisfying assignments and much less on the density of 
the problem. Both under-constrained and over-constrained problems with 
a small number of satisfying assignments turn out to be hard for 
randomized algorithms. In the same time, Davis-Putnam procedure, while
sensitive to the density, is quite robust with respect to the number of
satisfying truth assignments.

On the other hand, there are classes of problems that are ``easy'' 
for Davis-Putnam
procedure. For instance, Davis-Putnam procedure is very effective in
finding 3-colorings of graphs from special classes such as 2-trees
(see Section \ref{number} for definitions). Thus, they are not appropriate
benchmarks for Davis-Putnam type algorithms. However, a common intuition
is that structured problems are ``hard'' for randomized algorithms
\cite{skc96,sk93,skc94}. In this
paper we study this claim for the formulas that encode 3- and
4-coloring problem for 2-trees. We show that GSAT's running time 
relative to a given accuracy grows exponentially with the size of 
a graph. This provides a formal evidence to the ``hardness'' claim for 
this class of problems and implies that, while not useful in the studies 
of complete algorithms such as Davis-Putnam method, they are excellent 
benchmarks for studying the performance of randomized algorithms. 

The main contribution of our paper is not as much a discovery of
an unexpected behavior of randomized algorithms for testing
satisfiability as it is a proposed methodology for studying them. Our 
concepts of the accuracy and the relative running time allow us to 
quantify claims that are often accepted on the basis of intuitive
arguments but have not been formally pinpointed. 

In the paper, we apply our approach to the algorithm RWS-GSAT
from \cite{sk93,skc94}. This algorithm is commonly regarded as one of
the best
randomized algorithms for satisfiability testing to date. For our
experiments we used walksat version 35  downloaded from
ftp.research.att.com/dist/ai and run on a SPARC Station 20.
\section{Accuracy and running time}
\label{prel}

In this section, we will formally introduce the notion of the accuracy
of a randomized algorithm $\cal A$. We will then define the concept of
the running time relative to accuracy. 

Let $\cal F$ be a finite set of {\em satisfiable} CNF formulas and let 
$\cal P$ be a probability distribution defined on $\cal F$. Let $\cal A$
be a sound algorithm (randomized or not) to test satisfiability. By the 
{\em accuracy} of $\cal A$ (relative to the probability space 
$({\cal F},{\cal P})$), we mean the probability that $\cal A$ finds 
a satisfying assignment for a formula generated from $\cal F$ according 
to the distribution $\cal P$. Clearly, the accuracy of complete
algorithms (for all possible spaces of satisfiable formulas) is 1 and,
intuitively, the higher the accuracy, the more ``complete'' is 
the algorithm for the space $({\cal F},{\cal P})$.

When studying and comparing randomized algorithms that are not complete, 
accuracy seems to be an important characteristics. It needs to be taken
into account --- in addition to the running time. Clearly, very fast
algorithms that often return no satisfying assignments, even if they
exist, are not satisfactory. In fact, most of the work
on developing better randomized algorithms can be viewed as aimed at 
increasing the accuracy of these algorithms. Despite this, the accuracy
is rarely explicitly mentioned and studied (see \cite{spe96,msg97}). 

We will propose now an approach through which the running times of
randomized satisfiability testing algorithms can be compared. We will
restrict our considerations to the class of randomized algorithms 
designed according to the following general pattern. These algorithms
consist of a series of {\em tries}. In each try, a truth assignment 
is randomly generated. This truth assignment is then subject to 
a series of local improvement steps aimed at, eventually, reaching 
a satisfying assignment. The maximum number of tries the algorithm 
will attempt and the length of each try are the parameters of 
the algorithm. They are usually specified by the user. We will denote 
by $MT$ the maximum number of tries and by $MF$ --- the maximum number 
of local improvement steps.  Algorithms designed according to this 
pattern differ, besides possible differences in the values $MT$ and $MF$, 
in the specific definition of the local improvement process. A class of
algorithms of this structure is quite wide and contains, in particular,
the GSAT family of algorithms, as well as algorithms based on 
the simulated annealing approach.

Let $A$ be a randomized algorithm falling into the class described above. 
Clearly, its average running time on instances from the space 
$({\cal F},{\cal P})$ of satisfiable formulas depends, to
a large degree, on the particular choices for $MT$ and $MF$. To get an
objective measure of the running time, independent of $MT$ and $MF$,
when defining time, we require that a postulated accuracy be met.
Formally, let $a$, $0 < a\leq 1$, be a real number (a postulated accuracy). 
Define the {\em running time of $A$ relative to accuracy $a$}, $t^a$, to be 
the minimum time $t$ such that for some positive integers $MT$ and $MF$,
the algorithm $A$ with the maximum of $MT$ tries and with the maximum of
$MF$ local improvement steps per try satisfies:
\begin{enumerate}
\item the average running time on instances from $({\cal F},{\cal P})$
is at most $t$, and
\item the accuracy of $A$ on $({\cal F},{\cal P})$ is at least $a$.
\end{enumerate}
Intuitively, $t^a$ is the minimum expected time that guarantees 
accuracy $a$. In Section \ref{rand}, we describe an experimental approach
that can be used to estimate the relative running time.

The concepts of accuracy and accuracy relative to the running time
open a number of important (and, undoubtedly, very difficult) theoretical
problems. However, in this paper we will focus on an experimental study 
of accuracy and relative running time for a GSAT-type algorithm.
These algorithms follow the following general pattern for the local
improvement process. Given a truth assignment, GSAT selects a
variable such that after its truth value is {\em flipped} (changed 
to the opposite one) the number of unsatisfied clauses is minimum. Then,
the flip is actually made depending on the result of some additional
(often again random) procedure.

In our experiments, we used two types of data sets.
Data sets of the first type consist of randomly generated 3-CNF formulas 
\cite{msl92}. Data sets of the second type
consist of CNF formulas encoding the $k$-colorability problem for
randomly generated 2-trees. These two classes of data sets, as well as
the results of the experiments, are described
in detail in the next two sections. 

\section{Random 3-CNF formulas}
\label{rand}

Consider a randomly generated 3-CNF formula $F$, with $N$ variables and 
the ratio of clauses to variables equal to $L$. Intuitively, when 
$L$ increases, the probability that $F$ is
satisfiable should decrease. It is indeed so \cite{msl92}. What is more
surprising, it switches from being close to one to being close to zero
very abruptly in a very small range from $L$ approximately $4.25$ to
$4.3$.  The set of 3-CNF formulas at
the {\em cross-over} region will be denoted by $CR(N)$. 
Implementations of the Davis-Putnam procedure take, on average, the most 
time on 3-CNF formulas generated (according to a uniform probability 
distribution) from the cross-over regions. Thus, these formulas are commonly 
regarded as good test cases for experimental studies of the performance 
of satisfiability algorithms \cite{ca93,fre96}.

We used seven sets of {\em satisfiable} 3-CNF formulas generated from the 
{\em cross-over} regions $CR(N)$, $N=100, 150, \ldots, 400$.
These data sets are denoted by $DS(N)$. Each data set $DS(N)$
was obtained by generating randomly 3-CNF formulas with $N$ variables and 
$L = 4.30$ (for $N=100$) and $L= 4.25$ (for $N\geq 150$) clauses. For each 
formula, the Davis-Putnam algorithm was then used to decide its 
satisfiability. The first one thousand satisfiable formulas found in this 
way were chosen to form the data set.

The random algorithms are often used with much larger values of $N$ than
we have reported in this paper.  The importance of accuracy in this
study required that we have only {\em satisfiable} formulas (otherwise,
the accuracy cannot be reliably estimated). This limited the
size of randomly generated 3-CNF formulas used in our study since we
had to use a complete satisfiability testing procedure to discard those
randomly generated formulas that were not satisfiable. In Section
\ref{col}, we discuss ways in which hard test cases for randomized algorithms
can be generated that are not subject to the size limitation.


For each data set $DS(N)$, we determined values for $MF$, say $MF_1,
\ldots, MF_m$ and $MT_1,\ldots,$ $MT_n$ for use with RWS-GSAT, big enough 
to result in the accuracy at least 0.98. For instance, for $N=100$, $MF$ ranged
from $100$ to $1000$, with the increment of 100, and $MT$ ranged from 5
to 50, with the increment of 5. Next, for each combination of $MF$ and
$MT$, we ran RWS-GSAT on all formulas in $DS(N)$ and tabulated both the
running time and the percentage of problems for which the satisfying
assignment was found (this quantity was used as an estimate of the
accuracy). These estimates and average running times
for the data set $DS(100)$ are shown in the tables in Figure \ref{tbl-1}.

\begin{figure}[h]
{\footnotesize \begin{center}
\begin{tabular}{|r||r|r|r|r|r|r|r|r|r|r|}
\hline
MT &\multicolumn{10}{|c|}{ RWS-GSAT N=100 L=4.3 (time in seconds)}\\
\hline
50 &0.07 &0.07 &0.06 &0.05 &0.04 &0.05 &0.05 &0.06 &0.05 &0.05  \\
45 &0.06 &0.06 &0.05 &0.05 &0.04 &0.04 &0.04 &0.05 &0.05 &0.04  \\
40 &0.05 &0.05 &0.05 &0.04 &0.04 &0.04 &0.05 &0.03 &0.04 &0.03  \\
35 &0.05 &0.05 &0.04 &0.05 &0.05 &0.04 &0.04 &0.05 &0.05 &0.04  \\
30 &0.04 &0.04 &0.04 &0.04 &0.05 &0.04 &0.04 &0.04 &0.04 &0.04  \\
25 &0.03 &0.04 &0.04 &0.04 &0.04 &0.04 &0.04 &0.04 &0.04 &0.04  \\
20 &0.03 &0.03 &0.03 &0.04 &0.04 &0.04 &0.04 &0.03 &0.04 &0.03  \\
15 &0.02 &0.03 &0.03 &0.03 &0.04 &0.03 &0.04 &0.04 &0.03 &0.04  \\
10 &0.02 &0.02 &0.02 &0.03 &0.03 &0.03 &0.03 &0.03 &0.03 &0.03  \\
5 &0.01 &0.01 &0.01 &0.02 &0.02 &0.02 &0.02 &0.02 &0.02 &0.02  \\
\hline
MF &100 &200 &300 &400 &500 &600 &700 &800 &900 &1000  \\
\hline \end{tabular} 
\end{center}

\begin{center}
\begin{tabular}{|r||r|r|r|r|r|r|r|r|r|r|}
\hline
MT &\multicolumn{10}{|c|}{ RWS-GSAT N=100 L=4.3 (accuracy)}\\
\hline
50 &26 &72 &84 &98 &99 &97 &98 &97 &99 &100  \\
45 &26 &70 &87 &90 &97 &96 &99 &99 &98 &100  \\
40 &23 &71 &81 &94 &98 &98 &98 &100 &100 &99  \\
35 &24 &61 &87 &89 &94 &98 &95 &96 &99 &100  \\
30 &17 &55 &81 &89 &91 &96 &96 &98 &97 &100  \\
25 &20 &61 &69 &89 &89 &94 &97 &96 &97 &98  \\
20 &12 &52 &76 &84 &90 &92 &96 &96 &96 &96  \\
15 &11 &52 &68 &76 &84 &88 &94 &95 &95 &94  \\
10 &7 &36 &54 &66 &76 &81 &89 &89 &88 &91  \\
5 &6 &17 &38 &50 &63 &70 &66 &84 &84 &81  \\
\hline
MF &100 &200 &300 &400 &500 &600 &700 &800 &900 &1000  \\
\hline \end{tabular} 
\end{center}

\caption{\footnotesize \bf Tables showing accuracy and running time for instances
with $N=100$, $L=4.3$
and parameters $MF=100,\dots,1000$ and $MT=5,10,\ldots,50$}
\label{tbl-1}}
\end{figure}

Fixing a required accuracy, say at a level of
$a$, we then looked for the best time which resulted in this (or
higher) accuracy. We used this time as an experimental estimate for
$t^a$. For instance, there are 12 entries in the accuracy table with
accuracy $0.99$ or more. The lowest value from the corresponding
entries in the running time table is 0.03 sec. and it is used as an
estimate for $t^{0.99}$.

The relative running times $t^a$ for RWS-GSAT run on the data sets $DS(N)$, 
$N=100,150,\ldots, 400$, and for $a=0.90$ and $a=0.95$, are shown in 
Figure \ref{fig-1}. Both graphs demonstrate exponential
growth, with the running time increasing by the factor of 1.5 - 2 for
every 50 additional variables in the input problems. Thus, while GSAT
outperforms Davis-Putnam procedure for instances generated from
the critical regions, if we prescribe the accuracy, it is still
exponential and, thus, will quickly reach the limits of its
applicability. We did not extend our results beyond formulas with up to
400 variables due to the limitations of the Davis-Putnam procedure,
(or any other complete method to test satisfiability). For problems of
this size, GSAT is still extremely effective (takes only about 2.5
seconds). Data sets used in Section \ref{col} do not have this
limitation (we know all formulas in these sets are satisfiable and there
is no need to refer to complete satisfiability testing programs). The
results presented there also illustrate the exponential growth of the
relative running time and are consistent with those discussed here.

{\small\bf
\begin{figure}[h]
\centerline{\hbox{\psfig{figure=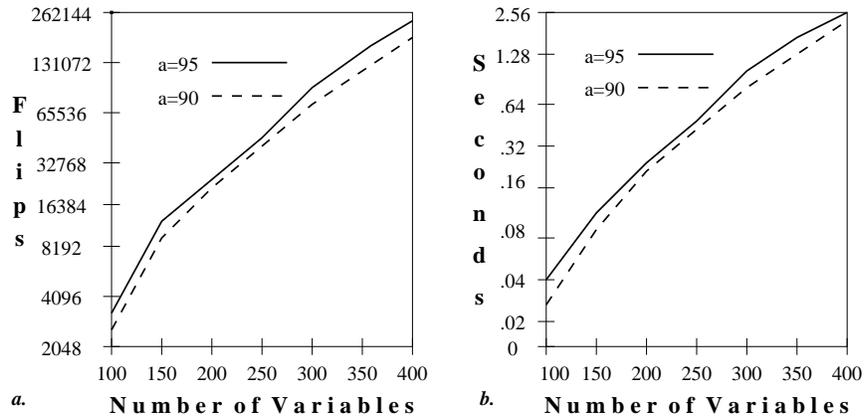}}}
\caption{\footnotesize \bf a) 
Average total flips of RWS-GSAT on randomly generated 3-CNF
formulas, plotted on a logarithmic scale as a function of 
the number of variables. This is included for a machine independent
comparison.
b) 
Running time of RWS-GSAT on randomly generated 3-CNF
formulas, plotted on a logarithmic scale as a function of 
the number of variables. }
\label{fig-1}
\end{figure}
}

\section{Number of satisfying assignments}
\label{number}

It seems intuitive that accuracy and running time would be
dependent on the number of possible satisfying assignments. 
Studies using randomly generated 3-CNF formulas \cite{cfgmtw96} and 
3-CNF formulas generated randomly with parameters allowing the user to
control the number of satisfiable solutions for each instance \cite{ci95} 
show this correlation.

In the same way as for the data sets $DS(N)$, we constructed data sets 
$DS(100,p_{k-1},p_k)$, where $p_0 = 1$, and $p_k = 2^{k-3}*100$, 
$k=2,\ldots,11$. Each data set $DS(100,p_{k-1},p_k)$ 
consists of 100 satisfiable 3-CNF formulas generated from the cross-over 
region $CR(100)$ and having more than $p_{k-1}$ and no more than $p_k$
satisfying assignments. Each data set was formed by randomly generating
3-CNF formulas from the cross-over region $CR(100)$ and by selecting
the first 100 formulas with the number of satisfying assignments
falling in the prescribed range (again, we used the Davis-Putnam
procedure, here).

For each data set we ran the RWS-GSAT algorithm with 
$MF = 500$ and $MT=50$ thus, allowing the same upper limits for 
the number of random steps for all data sets (these values
resulted in the accuracy of .99 in our experiments
with the data set $DS(100)$ discussed earlier).
Figure \ref{tab-1} summarizes our findings. It shows that there is 
a strong relationship between accuracy and the number of possible
satisfying assignments. 
Generally, instances with small number of solutions are much harder 
for RWS-GSAT than those with large numbers of solutions. Moreover, 
this observation is not affected by how 
constrained the input formulas are. We observed the same general
behavior when we repeated the experiment for data sets of 3-CNF formulas
generated from the under-constrained region (100 variables, 410 clauses) 
and over-constrained region (100 variables, 450 clauses), with
under-constrained instances with few solutions being the hardest. 

These
results indicate that, when generating data sets for experimental
studies of randomized algorithms, it is more important to ensure
that they have few solutions rather than that they come from the
critically constrained region.

{\small\bf
\begin{figure}[h]
\centerline{\hbox{\psfig{figure=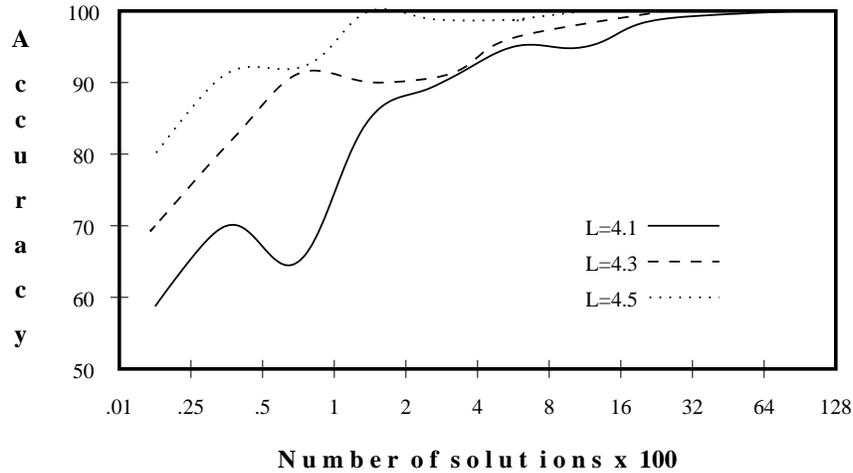}}}
\caption{\footnotesize \bf Accuracy of RWS-GSAT as a function of the number of
satisfying assignments}
\label{tab-1}
\end{figure}
}

\section{CNF formulas encoding $k$-colorability}
\label{col}

To expand the scope of applicability of our results and argue their
robustness, we also used in our study data sets consisting of CNF 
formulas encoding the $k$-colorability problem for graphs. While easy
for Davis-Putnam procedure (which resolves their satisfiability in
polynomial time), formulas of this type are believed to be ``hard'' for
randomized algorithms and were used in the past in the experimental 
studies of their performance. In particular, it was reported in 
\cite{sk93} that RWS-GSAT does not perform well on such inputs 
(see also \cite{jams91}). 

Given a graph $G$ with the vertex set $V=\{v_1,\ldots,v_n\}$ and the
edge set $E =\{e_1,\ldots, e_m\}$, we construct the CNF formula $COL(G,k)$
as follows. First, we introduce new propositional variables $col(v,i)$,
$v\in V$ and $i=1,\ldots,k$. The variable $col(v,i)$ expresses the fact
that the vertex $v$ is colored with the color $i$. Now, we define
$COL(G,k)$ to consist of the following clauses:
\begin{enumerate}
\item $\neg col(x,i) \vee \neg col(y,i)$, for every edge $\{x,y\}$ from
$G$,
\item $col(x,1)\vee \ldots \vee col(x,k)$, for every vertex $x$ of $G$,
\item $\neg col(x,i) \vee \neg col(x,j)$, for every vertex $x$ of $G$
and for every $i,j$, $1\leq i<j\leq k$. 
\end{enumerate}

It is easy to see that there is a one-to-one
correspondence between $k$-colorings of $G$ and satisfying assignments
for $COL(k,G)$. To generate formulas for experimenting with RWS-GSAT (and 
other satisfiability testing procedures) it is, then, enough to generate
graphs $G$ and produce formulas $COL(G,k)$.

\begin{figure}[h]
\centerline{\hbox{\psfig{figure=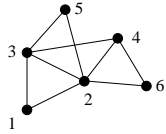}}}
\caption{\footnotesize \bf An example 2-tree with 6 vertices}
\label{fig-2}
\end{figure}

In our experiments, we used formulas that encode $3$-colorings
for graphs known as {\em $2$-trees}. The class of 2-trees is defined
inductively as follows:
\begin{enumerate}
\item A complete graph on three vertices (a ``triangle'') is a 2-tree
\item If $T$ is a 2-tree than a graph obtained by selecting an edge
$\{x,y\}$ in $T$, adding to $T$ a new vertex $z$ and joining $z$ to $x$
and $y$ is also a 2-tree.
\end{enumerate}
A 2-tree with 6 vertices is shown in Fig. \ref{fig-2}. The vertices
of the original triangle are labeled 1, 2 and 3. The remaining vertices
are labeled according to the order they were added.

The concept of 2-trees can be generalized to $k$-trees, for an arbitrary
$k\geq 2$. Graphs in these classes are important. They have bounded
tree-width and, consequently, many NP-complete problems can be solved
for them in polynomial time \cite{ap89}.


We can generate 2-trees randomly by simulating the definition given
above and by selecting an edge for ``expansion'' randomly in the current 
2-tree $T$. We generated in this way families $G(p)$,
for $p=50,60,\ldots, 150$, each consisting of one hundred
randomly generated 2-trees with $p$ vertices. Then, we created sets of
CNF formulas $C(p,3) = \{COL(T,3)\colon T\in G(p)\}$, for $p=50,60,
\ldots, 150$. Each formula in a set $C(p,3)$ has exactly 6
satisfying assignments (since each 2-tree has exactly 6 different
3-colorings). Thus, they are appropriate for testing the accuracy of
RWS-GSAT. 

Using CNF formulas of this type has an important benefit.
Data sets can be prepared without the need to use complete (but very
inefficient for large inputs) satisfiability testing procedures.
By appropriately choosing the underlying graphs, we can guarantee
the satisfiability of the resulting formulas and, often, we also have
some control over the number of solutions (for instance, in the case of
3-colorability of 2-trees there are exactly 6 solutions).

We used the same methodology as the one described in the previous
section to tabulate the accuracy and the running time of RSW-GSAT
for a large range of choices for the parameters $MF$ and $MT$.
Based on these tables, as before, we computed estimates for the times 
$t^a$ for $a= 0.95$, for each of the data sets. The results
that present the running time $t^a$ as a function of the number of
vertices in a graph (which is of the same order as the number of
variables in the corresponding CNF formula) are gathered in Figure
\ref{fig-3}. They show that RWS-GSAT's performance deteriorates
exponentially (time grows by the factor of $3-4$ for every 50 additional
vertices).


{\small\bf
\begin{figure}[h]
\centerline{\hbox{\psfig{figure=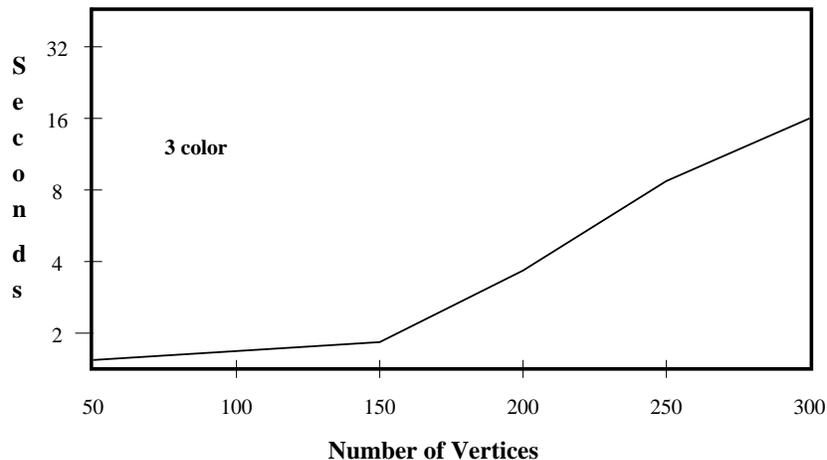}}}
\caption{\footnotesize \bf Running time of RWS-GSAT on formulas encoding 3-colorability,
plotted on a logarithmic scale as a function of the number of vertices.}
\label{fig-3}
\end{figure}
}

An important question is: how to approach constraint satisfaction
problems if they seem to be beyond the scope of applicability of
randomized algorithms? A common approach is to relax some constraints.
It often works because the resulting constraint sets (theories) are 
``easier'' to satisfy (admit more satisfying assignments). We have
already discussed the issue of the number of solutions in the previous 
section. Now, we will illustrate the effect of increasing the number 
of solutions (relaxing
the constraints) in the case of the colorability problem. To this end,
we will consider formulas from the spaces $C(p,4)$, representing
4-colorability of 2-trees. These formulas have exponentially many
satisfying truth assignments (a 2-tree with $p$ vertices has exactly
$3\times 2^{p}$ 4-colorings). For these formulas we also tabulated the
times $t^a$, for $a=0.95$, as a function of the number of vertices in the
graph. The results are shown in Figure \ref{fig-4}.

{\small\bf
\begin{figure}[h]
\centerline{\hbox{\psfig{figure=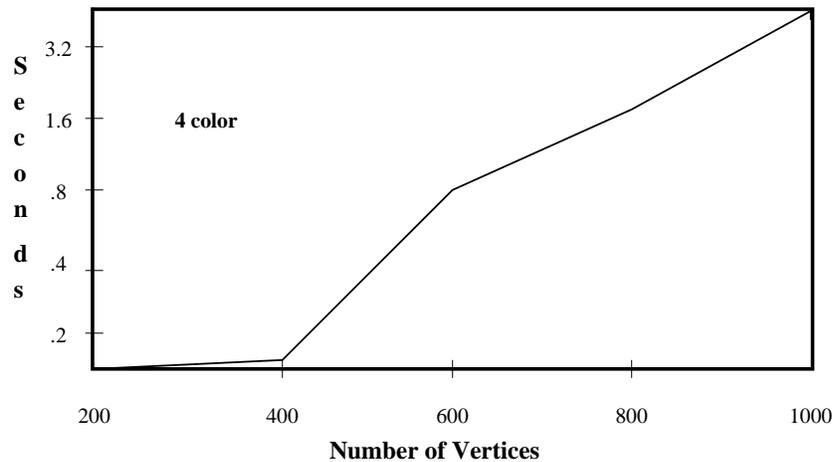}}}
\caption{\footnotesize \bf Running time of RWS-GSAT on formulas encoding 4-colorability, 
plotted on a logarithmic scale as a function of the number of vertices.}
\label{fig-4}
\end{figure}
}

Thus, {\em despite} the fact the size of a formula from $C(p,4)$ is 
larger than the size of a formula from $C(p,3)$ by the factor of $\approx 1.6$,
RWS-GSAT's running times are much lower. In particular, within .5 seconds 
RWS-GSAT can find a 4-coloring of randomly generated 2-trees with 500 vertices.
As demonstrated by Figure \ref{fig-3}, RWS-GSAT would require thousands
of seconds for 2-trees of this size to guarantee the same accuracy when 
finding 3-colorings. Thus, even a rather modest relaxation of constraints
can increase the number of satisfying assignments substantially enough
to lead to noticeable speed-ups. On the other hand, even though
``easier'', the theories encoding the 4-colorability problem 
for 2-trees still are hard to solve by GSAT as the rate of growth of the
relative running time is exponential (Fig. \ref{fig-4}).

The results of this section further confirm and provide quantitative
insights into our earlier claims about the exponential behavior of 
the relative running time for GSAT and on the dependence of the relative 
running time on the number of solutions. However, they also point out that 
by selecting a class of graphs (we selected the class of 2-trees here 
but there are, clearly, many other possibilities) and a graph problem 
(we focused on colorability but there are many other problems such as 
hamiltonicity, existence of vertex covers, cliques, etc.) then encoding 
these problems for graphs from the selected class yields a family of 
formulas that can be used in testing satisfiability algorithms. The main 
benefit of the approach is that by selecting a suitable class of graphs, 
we can guarantee satisfiability of the resulting formulas and can control 
the number of solutions, thus eliminating the need to resort to complete 
satisfiability procedures when preparing the test cases. We intend to
further pursue this direction.

\section{Conclusions}

In the paper we formally stated the definitions of the accuracy of a
randomized algorithm and of its running time relative to a prescribed 
accuracy. We showed that these notions enable objective studies and
comparisons of the performance and quality of randomized algorithms.
We applied our approach to study the RSW-GSAT algorithm. We showed that, 
given a prescribed accuracy, the running time of RWS-GSAT was exponential 
in the number of variables for several classes of randomly generated CNF 
formulas. We also showed that the accuracy (and, consequently, the running 
time relative to the accuracy) strongly depended on the number of satisfying 
assignments: the bigger this number, the easier was the problem for RWS-GSAT. 
This observation is independent of the ``density'' of the input
formula. The results suggest that satisfiable CNF formulas with few
satisfying assignments are hard for RWS-GSAT and should be used for
comparisons and benchmarking. One such class of formulas, CNF encodings
of the 3-colorability problem for 2-trees was described in the paper and
used in our study of RWS-GSAT.

Exponential behavior of RWS-GSAT points to the limitations of randomized
algorithms. However, our results indicating that 
input formulas with more solutions are ``easier'' for RWS-GSAT to deal 
with, explain RWS-GSAT's success in solving some large practical problems. 
They can be made ``easy'' for RWS-GSAT by relaxing some of the constraints.

\newcommand{\etalchar}[1]{$^{#1}$}

\end{document}